\documentclass{article}

\usepackage[english]{babel}

\usepackage[letterpaper,top=2cm,bottom=2cm,left=4cm,right=4cm,marginparwidth=2cm]{geometry}
\usepackage{authblk}

\usepackage{amsmath}
\usepackage{graphicx}
\usepackage[
backend=biber,
style=alphabetic,
sorting=ynt
]{biblatex}
\usepackage[colorlinks=true, allcolors=blue]{hyperref}

\title{NLP Research and Resources at DaSciM, Ecole Polytechnique}
\author[1]{Hadi ABDINE}
\author[1]{Yanzhu GUO}
\author[1]{Moussa KAMAL EDDINE}
\author[1]{Giannis NIKOLENTZOS}
\author[2]{Stamatis OUTSIOS}
\author[3]{Guokan SHANG}
\author[1]{Christos XYPOLOPOULOS}
\author[1]{Michalis VAZIRGIANNIS}
\affil[1]{Ecole Polytechnique, Institut Polytechnique de Paris, LIX}
\affil[2]{Athens University of Economics and Business, DB-NET}
\affil[3]{Linagora, LABS}

\begin{document}
\maketitle

\begin{abstract}
DaSciM (Data Science and Mining) part of LIX at Ecole Polytechnique, established in 2013 and since then producing research results in the area of large scale data analysis via methods of machine and deep learning. The group has been specifically active in the area of NLP and text mining with interesting results at methodological and resources level. Here follow our different contributions of interest to the AFIA community.
\end{abstract}

\section{Graph based representations for NLP and Text Mining} 

In recent years, graphs have become a widely used tool for modeling structured data. To enable the application of graph-based approaches to textual data, members of the DaSciM team developed the graph-of-words approach \cite{rousseau2013graph} which maps text to a graph where vertices correspond to terms and edges represent co-occurrences between the connected terms within  a fixed-size window. Once a document is represented as a graph, traditional, but also modern algorithms designed for graph-structured data can be applied to natural language texts. The researchers of the DaSciM team have explored how different text mining tasks can benefit from graph-based algorithms. One such task is \textit{keyword extraction}. Capitalizing on the concept of graph degeneracy, the $k$-core algorithm was applied to the graph representation of text to identify cohesive subgraphs \cite{rousseau2015main,tixier2016graph}. The vertices of these subgraphs can be considered as the most important terms (i.e., keywords) of a given textual document. The members of the DaSciM team have also utilized machine learning algorithms that operate on graphs to deal with tasks such as \textit{text categorization} and \textit{sentiment analysis}. Both graph kernels \cite{nikolentzos2017shortest} and graph neural networks \cite{nikolentzos2020message}, the two dominant methodologies for performing machine learning on graphs, have been applied to these problems with great success.
The effectiveness of the graph-based representations has also been evaluated in the task of \textit{detecting sub-events} from data collected from Twitter \cite{meladianos2015degeneracy,meladianos2018optimization}. The occurrence of a sub-event is usually associated with a change in the content of the messages posted recently by users compared to the content of the messages posted in the past. Such a significant change of content is reflected in the structure of the graph representation of tweets and can be captured by graph-based approaches.

\section{Word Senses and Disambiguation}
The number of senses of a given word, or polysemy, is a very subjective notion, which varies widely across annotators and resources. Creating high-quality, consistent, word sense inventories is a critical pre-requisite to successful word sense disambiguation. In \cite{xypolopoulos2020unsupervised} the DaSciM researchers propose a novel, fully unsupervised, and data-driven approach to quantify polysemy, based on basic geometry in the contextual embedding space. The proposed approach is based on multiresolution grids in the contextual embedding space. Such fully data-driven rankings of words according to polysemy can help in creating new sense inventories, but also in validating and interpreting existing ones. Additionally, the unsupervised nature of the method makes it applicable to any language.

\section{Abstractive Summarizarion for documents and meetings}
Abstractive summarization is an important and challenging task, requiring diverse and complex natural language understanding and generation capabilities. A good summarization model needs to read, comprehend,and write well. Like most of NLP tasks, the current state of the art is based on pretrained Transformers \cite{vaswani2017attention}. 

Trained on gigantic amounts of raw data and with hundreds of GPUs, models based on the Transformer architecture \cite{vaswani2017attention}, such as GPT \cite{radford2018improving} and BERT \cite{devlin2018bert}, have set new state-of-the-art performance in every NLU task.
Moreover, users around the world can easily benefit from these improvements, by finetuning the publicly available pretrained models to their specific applications.
This also saves considerable amounts of time, resources and energy, compared with training models from scratch.

BART \cite{lewis2019bart} combined a BERT-like bidirectional encoder with a GPT-like forward decoder, and pretrained this seq2seq architecture as a denoising autoencoder with a more general formulation of the masked language modeling objectives of BERT.
Since not only BART's encoder but also its decoder is pretrained, BART excels on tasks involving text generation.

The aforementioned efforts have made great strides. However, most of the research and resources were dedicated to the English language, despite a few notable exceptions. We partly address this limitation by contributing BARThez\footnote{named after a legendary French goalkeeper, Fabien Barthez: \tiny\url{ https://en.wikipedia.org/wiki/Fabien_Barthez}}, the first pretrained seq2seq model for French trained by the DaSciM team.
BARThez \cite{eddine2020barthez}, based on BART, was pretrained on a very large monolingual French corpus from past research that we adapted to suit BART's specific perturbation schemes.

Yet, while summarization for traditional textual documents (e.g., news) is an extensively-studied topic, summarization of multi-party conversations \cite{carenini2011methods, murray2008using, shang:tel-03169877} remains a comparably emerging and under-developed research area, even if it has recently been gaining  attention.  This asymmetry is due in large part to the nature of multi-party conversation, which poses challenges not encountered with traditional text, but also a lack of data and appropriate evaluation metrics. Such problems drove the DaSciM researchers to develop novel methods that move well beyond the state of the art for the task of abstractive meeting summarization \cite{shang-etal-2018-unsupervised}, as well as related sub-tasks in the area of spoken language understating, such as abstractive community detection \cite{shang-etal-2020-energy} and dialogue act classification \cite{shang-etal-2020-speaker}, as stepping stones towards generating better summaries.

\section{Applications on Legal text}
A long-standing application of NLP to legal documents is information extraction and retrieval from judicial decisions. The interest in mining data from judgments can be explained by the critical role they play in the administration of justice in both common and civil law systems. In \cite{boniol2020performance} the member of the DaSciM team used NLP methods to extract information from judgments of the French Court of appeal. They constructed indicators about the difficulty of lawyers’ performance and cases by using network analysis techniques on lawyers’ networks and cases’ networks. The objective of this research is to use these indicators to guide laypersons when confronted with the legal systems and contribute to the decrease of the access-to-justice gap by reducing the asymmetry of information characterizing the legal market.

\section{Large scale linguistic resources}
\textbf{French resources:}
Distributed word representations are popularly used in many tasks in natural language processing, adding that pre-trained word vectors on huge text corpus achieved high performance in many different NLP tasks. In \cite{abdine2021evaluation} DaSciM Researchers produced multiple high quality static word vectors for the French language using Word2vec CBOW where two of them are trained on huge 33GB crawled French data by the DaSciM team and the others are trained on an already existing French corpus. We also evaluate the quality of our proposed word vectors and  the existing French word vectors on the French word analogy task. In addition, we do the evaluation on multiple real NLP tasks that show the important performance enhancement of the pre-trained word vectors compared to the existing and random ones. \\ 

In addition to the static word vectors we publicly release the first large-scale pretrained seq2seq model dedicated to the French language, BARThez \cite{eddine2020barthez}, featuring 165M parameters, and trained on 101 GB of text for 60 hours with 128 GPUs.
We evaluate BARThez on five discriminative tasks and two generative tasks, with automated and human evaluation, and show that BARThez is very competitive with the state of the art.
To address the lack of generative tasks in the existing FLUE benchmark, we put together a novel dataset for summarization in French, OrangeSum, that we publicly release\footnote{ \url{https://github.com/Tixierae/OrangeSum}} and analyze in this paper.
OrangeSum is more abstractive than traditional summarization datasets, and can be considered the French equivalent of XSum. \\

We also introduce BERTweetFR \cite{guo2021bertweetfr}, the first large-scale pre-trained language model for French tweets. As a valuable resource for social media data, tweets are often written in an informal tone and have their own set of characteristics compared to conventional sources. Domain-adaptive pre-training is proven to provide significant benefits in helping models encode the complexity of specific textual domains. While efforts on domain adaptation of large-scale language models to tweets have been made in English, there is no similar work in any other language. Our model is initialized using a general-domain French language model CamemBERT which follows the base architecture of BERT. The adaptive pre-training is performed on 8 V100 GPUs with a 16GB dataset of 226M French tweets, lasting approximately 8 days. Experiments show that BERTweetFR outperforms all previous general-domain French language models on three downstream Twitter NLP tasks of offensiveness identification, named entity recognition and semantic shift detection. The dataset used in the offensiveness detection task is first created and annotated by our team, filling in the gap of such analytic datasets in French. We make our model publicly available in the transformers library with the aim of promoting future research in analytic tasks for French tweets.\\

In addition, we introduce JuriBERT \cite{douka2021juribert}, a set of BERT models (tiny, mini, small and base) pre-trained from scratch on French legal-domain specific corpora.
JuriBERT models are pretrained on 6.3GB of legal french raw text from two different sources: the first dataset is crawled from Légifrance and the other one consists of anonymized court’s decisions and the Claimant’s pleadings from the Court of Cassation. The latter contains more than 100k long documents from different court cases.
JuriBERT models are pretrained using Nvidia GTX 1080Ti and evaluated on a legal specific downstream task which consists of assigning the court Claimant’s pleadings to a chamber and a section of the court. While $JuriBERT_{SMALL}$ outperforms the general-domain BERT models ($CamemBERT_{BASE}$ and $CamemBERT_{LARGE}$), the other models have a similar performance.\\

Finally, we created a web application\footnote{\url{http://nlp.polytechnique.fr/}} to test and visualize the quality of BARThez, BERTweetFR and the obtained static word embeddings.
The produced French word embeddings are available to the public, along with the fine-tuning code.\\

\textbf{Greek resources:}
In \cite{outsios2018word}, Greek Web was used to produce a large scale clear text corpus and then various resources
like trained vectors, stopwords, vocabulary, unigrams, bigrams and trigrams.
In \cite{outsios2019evaluation} we evaluated our newly trained vectors using two newly produced datasets: A Word analogy test set and a Word similarity data-set (WordSim353).
All our produced resources are publicly available\footnotemark.
\footnotetext{\url{http://archive.aueb.gr:7000/resources/}}



\printbibliography

\end{document}